\documentclass{article}

\usepackage{arxiv}

\usepackage[utf8]{inputenc} 
\usepackage[T1]{fontenc}    
\usepackage{hyperref}       
\usepackage{url}            
\usepackage{booktabs}       
\usepackage{amsfonts}       
\usepackage{nicefrac}       
\usepackage{microtype}      
\usepackage{adjustbox}
\usepackage{lipsum}
\usepackage[hypcap=false]{caption}
\usepackage{graphicx}
\usepackage{makecell}
\graphicspath{ {./images/} }
\usepackage[
backend=bibtex,
style=numeric,
sorting=none
]{biblatex}
\usepackage{xpatch}

\makeatletter         
\patchcmd\blx@bblinput{\blx@blxinit}
                      {\blx@blxinit
                       
                      }{}{\fail}
\makeatother

\hypersetup{
    colorlinks=true,
    linkcolor=red,
    pdftitle={BackgroundMatting, H. Javidnia et al.},
    citecolor=green,
    urlcolor=black
    }

\title{Background Matting}

\author{
Hossein Javidnia\\
ADAPT Centre\\
School of Computer Science and Statistics\\  
Trinity College Dublin, Ireland\\
\texttt{javidnih@tcd.ie}
\And
François Pitié\\
Sigmedia Group\\
Department of Electronic \& Electrical Engineering\\
Trinity College Dublin, Ireland\\
\texttt{pitief@tcd.ie}
}

\begin{document}
\maketitle

\begin{abstract}
The current state of the art alpha matting methods mainly rely on the trimap as the secondary and only guidance to estimate alpha. This paper investigates the effects of utilising the background information as well as trimap in the process of alpha calculation. To achieve this goal, a state of the art method, AlphaGan is adopted and modified to process the background information as an extra input channel. Extensive experiments are performed to analyse the effect of the background information in image and video matting such as training with mildly and heavily distorted backgrounds. Based on the quantitative evaluations performed on Adobe Composition-1k dataset, the proposed pipeline significantly outperforms the state of the art methods using AlphaMatting benchmark metrics.
\end{abstract}


\section{Introduction}
 Alpha estimation is a regression problem that calculates the opacity value of each blended pixel in the foreground object. It serves as a prerequisite for a broad range of applications such as movie post production, digital image editing and compositing live action.
 
 Formally, the composition image $I_i$ is represented as a linear combination of the background $B_i$ and foreground $F_i$ colors \cite{chuang2001bayesian}:
\begin{equation} \label{eq:1}
I_i = \alpha_iF_i+(1-\alpha_i)B_i
\end{equation}
where $\alpha_i \in [0,1]$ denotes the opacity or alpha matte of the foreground at pixel $i$.
Often, a user input is provided as a guidance in the form of a trimap, which assigns a label for every pixel as foreground $\alpha=1$, background $\alpha=0$ and unknown opacity. The goal of the matting algorithms is to estimate the unknown opacities by utilising the pixel color information of the known regions. Tackling the inverse problem of Eq. \ref{eq:1} is considerably difficult as there are 7 unknowns and 3 equations to be solved for an RGB image. The main motivation in this paper is to increase the matting accuracy by reducing the number of unknowns in Eq. \ref{eq:1}. To do so, we presume that the background information $B$, is known either by capturing a clear background or through reconstruction methods that can estimate the occluded background regions.

In traditional methods, the matte is estimated by inferring the alpha information in the unknown areas from those in known areas \cite{wang2008image}. For example, the matte values could be propagated from known to unknown areas based on the spatial and appearance affinity relation between them \cite{Aksoy2017matting,Chen2013matting,Chen2013mattlocal,He2010laplac,Lee2011nonlocal,Levin2008closedf,Levin2008spectral,Jian2004poisson}. An alternative solution is to compute the unknown mattes by sub-sampling the color and texture distribution of the foreground and background planes followed by an optimization such as likelihood of alpha values \cite{chuang2001bayesian,He2013iter,He2011global,Wang2005iter,Wang2007robust}. Despite the promising performance of these methods on public benchmarks, there is still an unresolved issue of natural image matting and consistency in videos between consecutive frames. One important reason causing this problem is the fact that the performance of these methods heavily rely on the accuracy of the given trimap. Generating the trimap for a sequence of images from a video is indeed a challenging task as it requires tracking the object of interest and defining an appropriate and relevant unknown areas to be solved.

To address these challenges, this paper presents a Background-Aware Generative Adversarial Network (AlphaGan-BG) which utilises the information present in the background plane to accurately estimate the alpha matte compensating for the issues caused by inaccurate trimap. Unlike the state of the art which only use RGB image and trimap as the input, AlphaGan-BG analyses the color and texture provided as background information to achieve a better accuracy.
To our best knowledge, this paper contributes the first deep learning approach which takes advantage of the background information to estimate alpha mattes. Both our qualitative and quantitative experiments demonstrate that AlphaGan-BG significantly outperforms the state of the art matting methods. 

\section{Previous Works}
Alpha matting is a well established and studied field of research in computer vision with a rich literature. A significant amount of work has been done over the past decade to address the issues in natural image matting. More recently, deep learning approaches have shown an impressive performance on various computer vision tasks including image matting too.

This section briefly reviews the state of the art alpha matting methods within two categories: conventional methods and deep learning based methods.

\subsection{Conventional Matting Methods}
The conventional alpha matting approaches could be categorised into sampling based and affinity based methods. Sampling based methods \cite{chuang2001bayesian,Gastal2010SharedSF,He2011global,Xiaoxue2016cluster} initially collect a set of known foreground and background color samples to identify the best foreground-background color pair for a pixel. 

The general rule is to use Eq. \ref{eq:1} to calculate the alpha value once the corresponding background and foreground colors are determined. The issue with the sampling based method is that they don't make use of the texture information present in the image and they don't enforce spatial smoothness thus introducing an additional spatial smoothness step. More importantly, there is always the ambiguity on how the samples are chosen and where are they chosen from; causing matte discontinuities.
For instance, Shared Matting \cite{Gastal2010SharedSF} select the samples from the trimap boundaries between the known and unknown pixels. Global Matting \cite{He2011global} makes use of all the pixels within the trimap boundary therefore increasing the performance time. Sparse sampling \cite{Karacan2015divergance} applies the sampling in a super-pixel level by assessing their similarity using KL-Divergence based distance measure.

Affinity-based methods work by analysing the affinities of neighboring pixels to propagate alpha information from known to unknown regions. Levin et al. \cite{Levin2008closedf} proposed a closed-form matting solution where the local color information is used to compute the affinity between two pixels. In \cite{Levin2008closedf} the alpha matte is calculated by solving a sparse linear system. The advantage of the closed-form solution is the prediction of the properties of the solution by analysing the eigenvectors of a sparse matrix. Chen et al. \cite{Xiaowu2012manifold} proposed a locally linear embedding system which represents every unknown pixel as a linear combination of its  neighbors. KNN matting \cite{Chen2013matting} utilised nonlocal principal to find the affinities. The basis of this principal is that a pixel is a weighted sum of the pixels with similar appearance to the given weight \cite{Chen2013matting}. This method enforces the the pixels and their corresponding nonlocal neighbors to have close alpha value. Aksoy et al. \cite{Aksoy2017matting} constructed their method based on color-mixture flow using pixel-to-pixel connections between the image and it's corresponding trimap. The flow is based on local linear embedding with gradual improvement in matting quality as more building blocks are added to the information flow. It was shown in \cite{Aksoy2017matting} that combining local and non-local affinities can result in a higher quality alpha matte. Several other state of the art approaches such as Random Walks \cite{grady2005random}, FuzzyMatte \cite{Zheng2008fuzzy}, Spectral Matting \cite{Levin2008spectral} and Geodesic Matting \cite{Bai2007geo} can also be categorised as affinity based methods. 

\subsection{Deep Learning Based Matting Methods}
Emerging field of deep learning along with the new generation of hardware, enabled many researches to tackle the issues of natural image matting with promising performances. Cho et al. \cite{Donghyeon2016DCNN} proposed an end-to-end Deep Convolutional Neural Networks (CNN) which utilises the results of the closed form matting and KNN matting for alpha estimation. Xu et al. \cite{Xu2017deepmatting} proposed a two part structure to predict alpha. The first part is an encoder-decoder module trained to predict the alpha from the input image and trimap; the second part is a small CNN trained to perform a post-processing step to increase the quality of the estimated alpha. Lu et al. \cite{Lu2019ICCV} proposed IndexNet Matting by introducing indexed pooling and upsampling operators. They modeled the indices as a function of the feature map to perform the upsampling. There are many other methods proposed to use deep learning to tackle the issues of natural image matting such as VDRN Matting \cite{Tang2019VDRN}, SampleNet Matting \cite{Tang2019CVPR}, AdaMatting \cite{Cai2019ICCV}, Late Fusion Matting \cite{Zhang2019CVPRlatefusion}, Inductive Guided Filter Matting \cite{Li2019inductive}, however, the analysis of these methods goes beyond the scope of our work.

\section{AlphaGan-BG Network}
\label{AlphaGanbgnet}
The framework in this research is built on the first proposed GAN to estimate alpha mattes. AlphaGAN \cite{Lutz2018alphagan} was introduced in 2018 motivated by the encoder-decoder structure proposed in \cite{Xu2017deepmatting}. The original architecture of AlphaGAN consists of a generator $G$ and discriminator $D$. 

In the original form of AlphaGAN, $G$ accepts the input in a form of a 4 channel volume made of a composited image (3 channels) and the corresponding trimap (1 channel). $D$ is responsible for distinguishing the real from fake input volume. The first 3 channels of the input volume to $D$ belongs to the RGB values of the new composited images based on predicted alpha and the last channel is the original trimap to help $D$ focus on salient regions.

AlphaGAN followed the same path as the rest of the state of the art methods with the assumption that the only data available is an RGB image and the corresponding trimap. However, in this paper, background information is also considered as the known variable and the input to the network.  

\subsection{Generator \texorpdfstring{$G$}{}}
In this research, $G$ is an encoder-decoder network that accepts the input in a form of a 7 channel volume, where the first 3 channels contain the RGB image, the second 3 channels contain the RGB background information and the last channel contains the trimap.
The encoder is based on ResNet50 \cite{He2016resnet} architecture pretrained on ImageNet \cite{Russakovsky2015imagenet} where the convolutions in the 3\textsuperscript{rd} and 4\textsuperscript{th} block of the ResNet are replaced by dilated convolutions with rate 2 and 4, respectively. To resample features at several scales, Atrous Spatial Pyramid Pooling (ASPP) module \cite{Chen2017ASPP,Chen2018ASPP} is added after ResNet block 4.

Similary to AlphaGAN, the decoder is simply a set of convolutional layers and skip connections from the encoder. The output of the encoder is bilinearly upsampled with the factor of 2 to maintain the same spatial resolution for the feature maps as the output of ResNet block 1. To reduce the dimensions, the output of the ResNet block 1 is fed into $1\times1$ convolutional layer and concatenated with the upsampled feature maps from encoder. This is followed by $3\times3$ convolutions and upsampling using the saved pooling indices in the first layer of the encoder. The results are once again concatenated with the feature maps from the encoder with the same resolution. Before feeding the output to the final set of convolution layers, transposed convolutions are applied to upsample it followed by a concatenation with the RGB input image. 
ReLU \cite{Vinod2010relu} activation functions and Batch Normalization \cite{Sergey2015batchnorm} layers are used for all the layers except the last one which utilises a sigmoid activation to scale the output between 0 and 1. Fig. \ref{netstruct} illustrates the encoder-decoder structure of $G$.
\begin{center}
\includegraphics[width=\columnwidth]{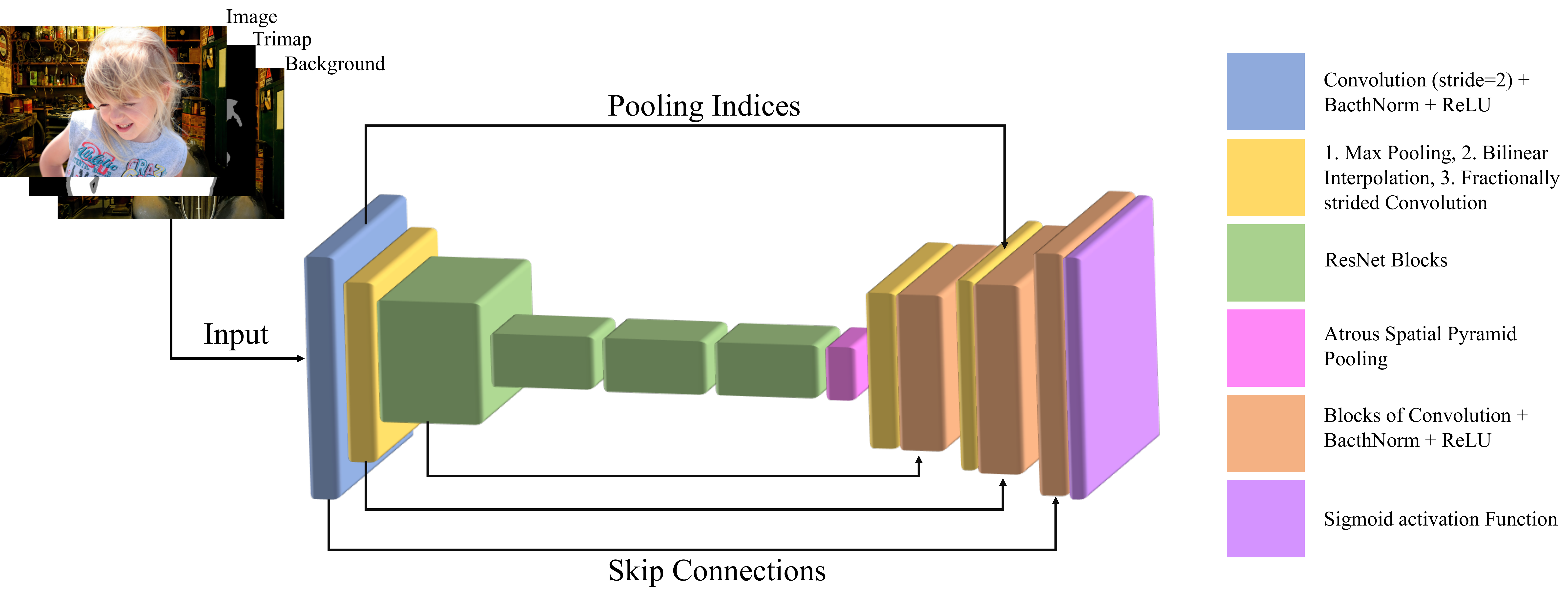}
\captionof{figure}{AlphaGan-BG: Structure of Generator ($G$).}\label{netstruct}
\end{center}

\subsection{Discriminator \texorpdfstring{$D$}{}}
This architecture employs PatchGAN \cite{Isola2017patchgan} as the discriminator. $D$ attempts to distinguish fake from real input which is a 7 channel volume. The real input is constructed by original composition using truth alpha, background and trimap. The fake input contains the new composition using the alpha generated by $G$, background and the trimap. By providing the background information, $D$ will enforce $G$ to output sharper and more accurate result as the issue of differentiating foreground and background is resolved by default. 

\subsection{Loss Functions}
The full objective of the network is a combination of three loss functions: alpha-prediction loss $\mathcal{L}_{alpha}$, compositional loss $\mathcal{L}_{comp}$ and adversarial loss $\mathcal{L}_{GAN}$ \cite{Goodfellow2014loss}:
\begin{equation} \label{eq:2}
\mathcal{L}_{total} = \mathcal{L}_{alpha}+\mathcal{L}_{comp}+\mathcal{L}_{GAN}
\end{equation}
$\mathcal{L}_{alpha}$ is the absolute difference of the ground truth and predicted alpha values for all the pixels.  $\mathcal{L}_{comp}$ is the absolute difference of the composited image using ground truth alpha and the composited image using predicted alpha. The composition in both cases are based on the ground truth foreground and background images \cite{Xu2017deepmatting}. $\mathcal{L}_{GAN}$ is defined based on the fundamentals of adversarial networks, where in this research, $G$ aims at generating alpha mattes close to the ground truth while $D$ aims at distinguishing real from fake input; resulting in $G$ minimizing the $\mathcal{L}_{GAN}$.

\section{Experiments and Discussion}
\subsection{Dataset}
The network in this paper is trained on Adobe Matting dataset \cite{Xu2017deepmatting} consists of 431 foreground images for training and 50 images for testing with corresponding ground truth. To augment the data, Pascal VOC 2008 \cite{everingham2010pascal} and MSCOCO images \cite{lin2014microsoft} are used as the background for image composition resulting in a training set containing 43100 images.

\subsection{Data Preparation}
\label{dataprep}
As described in Section \ref{AlphaGanbgnet}, this network takes advantage of the background information to predict alpha matte. This requires the background information to be available during the test phase as well as training. However, acquiring the background image during the test phase is a challenging task. To achieve this, several inpainting and background reconstruction methods \cite{Kim2019deepinpainting,Xu2019flowinpainting,Huang2016tempinpainting,LAUGRAUD201712,Laugraud2015medianbasedinpainting,Herling2014pixmix} are studied to analyse their accuracy and performance on static images and videos. The findings indicate that currently there is no background reconstruction method that can generate a clear background without artifacts. The ultimate goal is to obtain a reconstructed image which is equivalent of the original input used for the composition. The common artifacts present in the output of the reconstruction methods are the blur (degraded quality) and shift (translation), meaning that the region containing the object of interest is slightly translated in the reconstructed image.

To simulate these artifacts, two sets of backgrounds are augmented. In the first set, a random selection of images are manipulated by applying a hexagonal shape Gaussian blur with a random filter size. The location of the hexagonal blur is randomly chosen along the dimensions of the input image. The diameter of the shape is randomly selected between 120 and 345 pixels with rotation angle chosen by generating a linearly spaced vector. The blurred region is also translated using a 2D linear translation. In the second set, all the images are initially blurred followed by applying the hexagonal shape Gaussian blur at a random location.
Comparatively, the first scenario represents a more realistic case as it contains both clean and partially distorted backgrounds. However, the second set represents severely distorted cases where all the images are blurred with an additional distorted patch introducing a more challenging set for training.

\subsection{Training}
In this paper, two models are trained for evaluation purposes. The first model utilises the first set of background images as described in Section \ref{dataprep} and the second model uses the second set of backgrounds with severe distortion. In order to make the remaining sections easier to follow, we refer to the first model as \textit{AlphaGan-BG\_M} (Mildly distorted) and second model as \textit{AlphaGan-BG\_H} (Heavily distorted).

AlphaGan-BG\_M and AlphaGan-BG\_H are trained for 376 and 650 epochs respectively with the initial learning rate set to 0.0002. Adam optimizer \cite{kingma2014adam} with $\beta$ = 0.999 is also employed for optimization purposes.

\subsection{Results and Evaluation}
\subsection{Still Image Matting}
The evaluation for still images is performed on a set of 50 images from Adobe Matting \cite{Xu2017deepmatting} test set. Note that, none of the test images are considered as part of the training. Four metrics based on AlphaMatting benchmark \cite{Rhemann2009matting,alphamattingweb} are used for evaluation purposes including Sum of Absolute Difference (SAD), Mean Square Error (MSE), Connectivity (CONN) and Gradient Errors (GRAD). The test images from AlphaMatting benchmark are not considered as part of this evaluation as there is no available background information for the test set.
The background images used for evaluation are also manipulated using the pipeline describes in Section \ref{dataprep} to simulate the reconstruction artifacts. The performance of the trained models are compared against 8 state of the art methods ranked in AlphaMatting benchmark with publicly available code including Closed-Form Matting \cite{Levin2008closedf}, DCNN Matting \cite{Donghyeon2016DCNN}, Deep Matting \cite{Xu2017deepmatting}, IndexNet Matting \cite{Lu2019ICCV}, Information-flow Matting \cite{Aksoy2017matting}, KNN Matting \cite{Chen2013matting}, Late Fusion \cite{Zhang2019CVPRlatefusion} and AlphaGAN \cite{Lutz2018alphagan}.

\begin{center}
\begin{tabular}{l c c c c} \toprule
    {Methods} & {SAD} & {MSE} & {GRAD} & {CONN} \\ \midrule
    Closed-Form Matting \cite{Levin2008closedf}  &  78.768 & 0.065 & 57.047 & 56.856 \\
    DCNN Matting \cite{Donghyeon2016DCNN}  & 85.842 & 0.070 & 57.622  & 65.196 \\
    Deep Matting \cite{Xu2017deepmatting}  & 33.075 & 0.017 & 23.316  & 34.204 \\
   IndexNet Matting \cite{Lu2019ICCV}  & 28.984 & 0.013 & 19.127 & 28.872\\ 
    Information-flow Matting \cite{Aksoy2017matting}  & 84.766 & 0.067 & 52.789 & 63.827\\ 
    KNN Matting \cite{Chen2013matting}  & 95.122 & 0.082 & 66.188 & 74.940\\ 
    Late Fusion \cite{Zhang2019CVPRlatefusion}  & 88.109 & 0.097 & 59.382 & 91.743\\ 
    AlphaGAN \cite{Lutz2018alphagan}  & 35.057 & 0.019 & 33.598 & 35.963\\ \midrule
    AlphaGan-BG\_M  & \textbf{11.312} & \textbf{0.002} & \textbf{4.850} &  \textbf{8.696} \\
    AlphaGan-BG\_H & 14.692 & 0.003 & 8.410 & 12.328\\ \bottomrule
\end{tabular}
\captionof{table}{The quantitative comparison of the AlphaGan-BG models against state of the art. The best average value/metric is emboldened.}\label{tabcomparison}
\end{center}

\begin{center}
\includegraphics[width=\columnwidth]{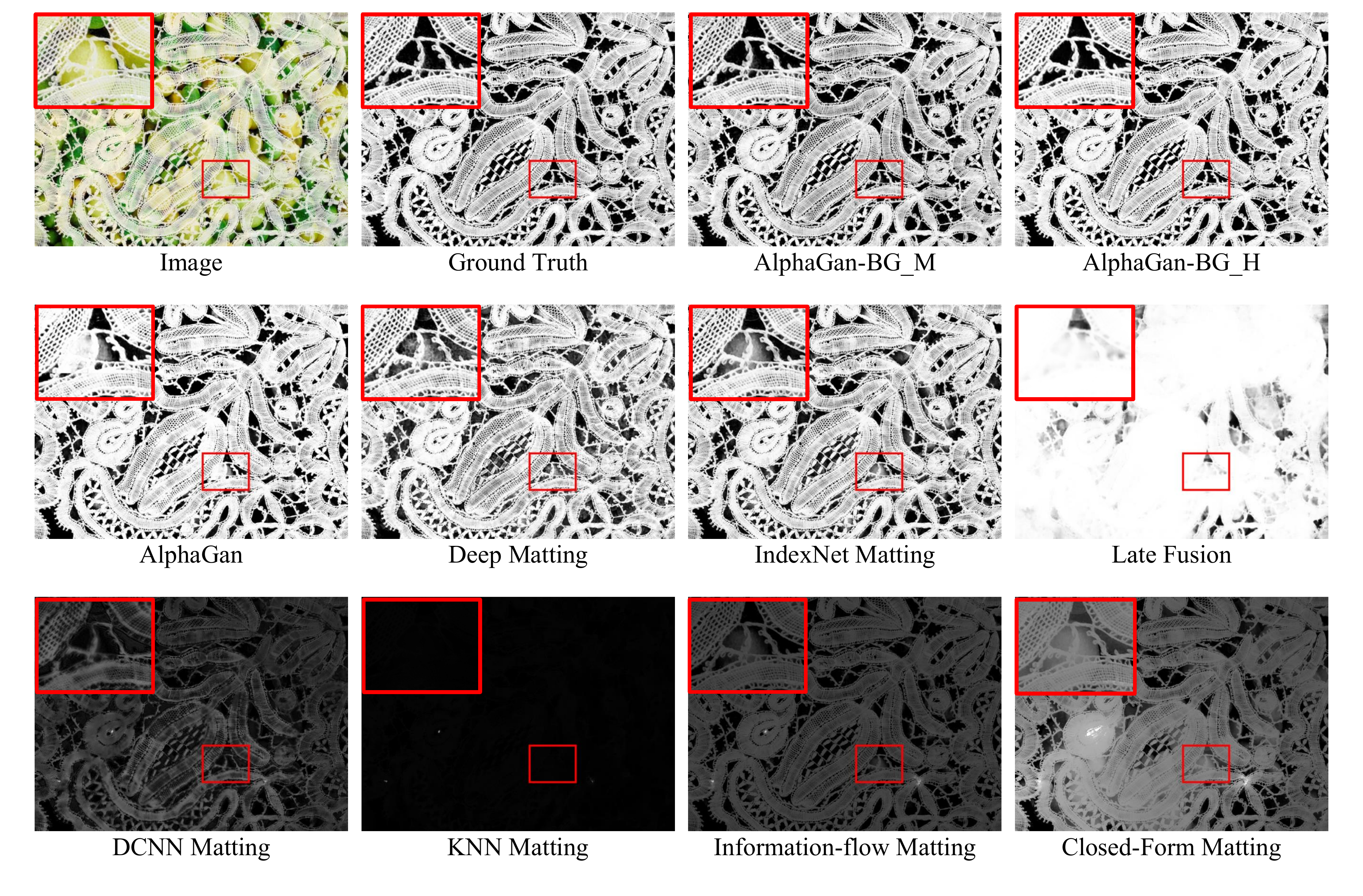}
\captionof{figure}{Comparison with State of the Art Methods - Example 1.}\label{compare1}
\end{center}
Table \ref{tabcomparison} presents the numerical evaluation of the AlphaGan-BG models against the state of the art methods and clearly notes that AlphaGan-BG outperforms the other methods based on the commonly used AlphaMatting benchmark metrics. This experiment also validates the idea of using background information for alpha estimation. As discussed in Section \ref{dataprep}, based on the current state of the art in background reconstruction, it is very challenging to obtain a clear reconstructed background; However, this experiment demonstrates that even having a partial information about the background plane (with distortion) can significantly increase the accuracy of the alpha prediction.

Fig. \ref{compare1} and Fig. \ref{compare2} illustrate the qualitative comparison of the proposed models against the state of the art methods. A part of the predicted alpha mattes by the state of the art is marked in Fig. \ref{compare1} to closely expose the difference in the performance of the methods.

\begin{center}
\includegraphics[width=\columnwidth]{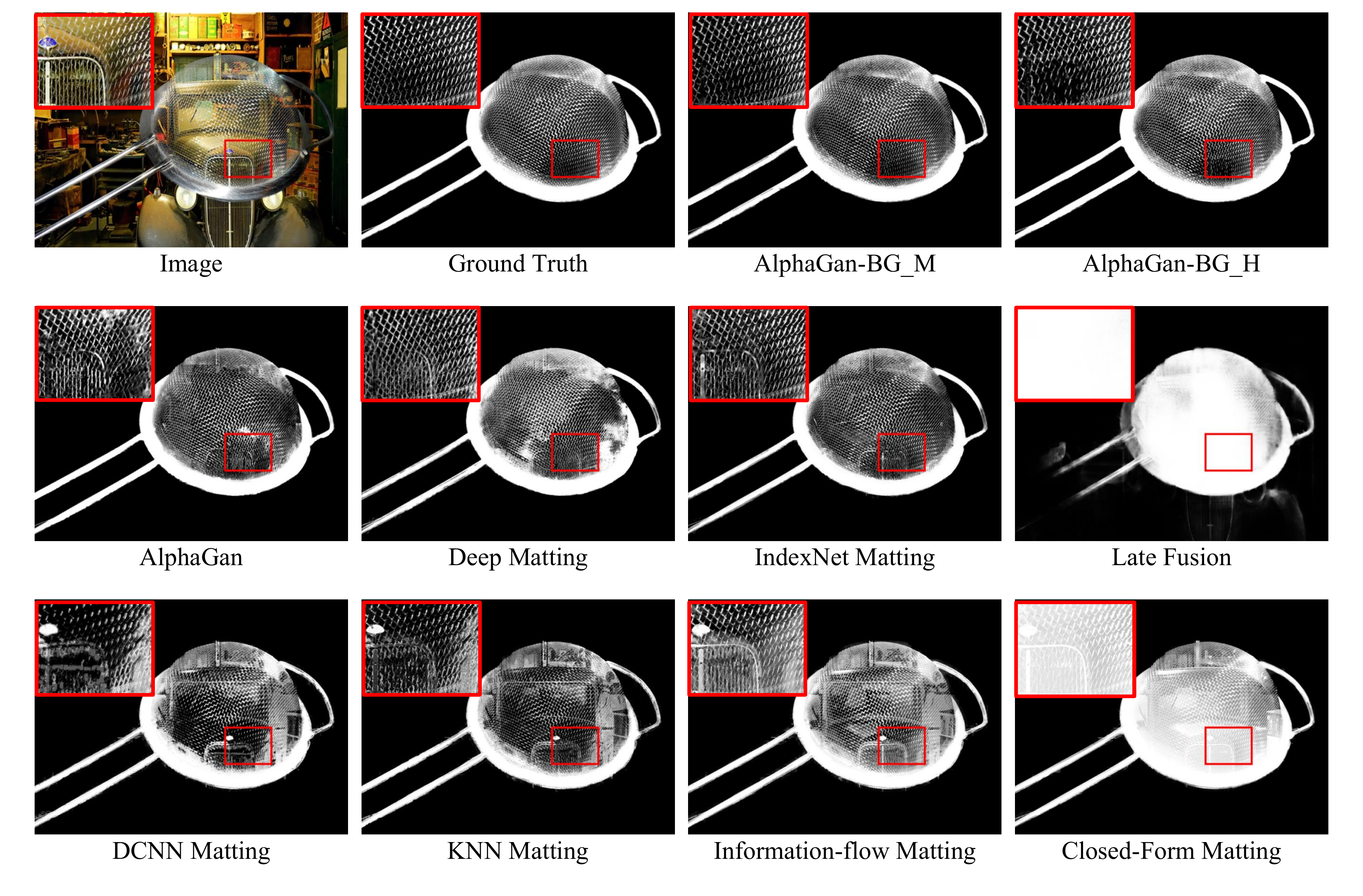}
\captionof{figure}{Comparison with State of the Art Methods - Example 2.}\label{compare2}
\end{center}
The performance of the AlphaGan-BG\_M and AlphaGan-BG\_H in Fig. \ref{compare1} is a clear example and proof of an earlier statement that including the partial background information of the image during the matting pipeline, can significantly increase its accuracy and preserve fine details.

Fig. \ref{compare2} is another example of the visual comparison against the state of the art where the superior performance of the AlphaGan-BG\_M and AlphaGan-BG\_H is clearly visible through the marked areas. For more and detailed visual results refer to Appendix 1.

\subsection{Video Matting}
To evaluate the performance of the AlphaGan-BG\_M and AlphaGan-BG\_H on video sequences, we used four state of the art background reconstruction and inpainting methods including Deep Video Inpainting (DVI) \cite{Kim2019deepinpainting}, Deep Flow Inpainting (DFGI) \cite{Xu2019flowinpainting}, FVC \cite{Afifi2014fastvideocomp} and Gated Video Inpainting (GVI) \cite{chang2019free} to separate the foreground and background layers. We also considered backgrounds with simulated artifacts as part of this evaluation. The background layers are further used as the input to the proposed matting framework.
Three video sequences including \textit{Alex}, \textit{Castle} and \textit{Dmitriy} from VideoMatting benchmark \cite{Erofeev2015videomatting} are used for evaluation purposes. Tables \ref{tabcomparisonvideosad}-\ref{tabcomparisonvideoconn} present the numerical evaluation of the AlphaGan-BG models on video sequences. The aforementioned reconstruction methods are applied to each sequence to extract the background layer as one of the input channels to AlphaGan-BG models. One important and obvious take from this experiment is the fact that a successful background aware matting method significantly relies on the quality of the reconstructed background. Although, the point of this experiment is not to compare the performance of the reconstruction methods, a few state of the art techniques such as FCV \cite{Afifi2014fastvideocomp} generate background layers with less artifacts and similar to the simulated ones resulting in more accurate alpha estimation using AlphaGan-BG\_M. On the other hand, AlphaGan-BG\_H performs better in scenarios where the reconstructed background layers are heavily distorted such as DVI \cite{Kim2019deepinpainting} and DFGI \cite{Xu2019flowinpainting}.
A detailed set of visual results for this section is provided in Appendix 2.

\begin{center}
\begin{adjustbox}{width=\textwidth}
\begin{tabular}{l|c|c|c|c|c|c|c|c|c|c|c|c|c|c|c}
\Xhline{2\arrayrulewidth}
 &
  \multicolumn{3}{c|}{W\textbackslash Artifact} &
  \multicolumn{3}{c|}{DVI \cite{Kim2019deepinpainting}} &
  \multicolumn{3}{c|}{DFGI \cite{Xu2019flowinpainting}} &
  \multicolumn{3}{c|}{GVI \cite{chang2019free}} &
  \multicolumn{3}{c}{FVC \cite{Afifi2014fastvideocomp}} \\ \hline
               & A & C & D & A & C & D & A & C & D & A & C & D & A & C & D \\ \Xhline{2\arrayrulewidth}
AlphaGan-BG\_M &\textbf{1.004}   &\textbf{10.145}   &\textbf{1.66}   &9.95   &95.712   &11.856   &1.787   &32.808   &2.28   &16.814   &89.4   &15.046   &1.165   &11.385   &1.781   \\
AlphaGan-BG\_H &\textbf{1.28}   &\textbf{28.062}   &\textbf{1.758}   &1.658   &53.513   &2.115   &1.292   &37.207   &1.775   &2.409   &56.7   &2.91   &1.3   &28.352   &1.77   \\ \Xhline{2\arrayrulewidth}
\end{tabular}
\end{adjustbox}
\captionof{table}{SAD Metric - Performance of the AlphaGan-BG models using different background reconstruction methods. \textit{A: Alex, C: Castle} and \textit{D: Dimitriy}. The best average value per model for each animation across all reconstruction method is emboldened.}\label{tabcomparisonvideosad}
\end{center}

\begin{center}
\begin{adjustbox}{width=\textwidth}
\begin{tabular}{l|c|c|c|c|c|c|c|c|c|c|c|c|c|c|c}
\Xhline{2\arrayrulewidth}
 &
  \multicolumn{3}{c|}{W\textbackslash Artifact} &
  \multicolumn{3}{c|}{DVI \cite{Kim2019deepinpainting}} &
  \multicolumn{3}{c|}{DFGI \cite{Xu2019flowinpainting}} &
  \multicolumn{3}{c|}{GVI \cite{chang2019free}} &
  \multicolumn{3}{c}{FVC \cite{Afifi2014fastvideocomp}} \\ \hline
               & A & C & D & A & C & D & A & C & D & A & C & D & A & C & D \\ \Xhline{2\arrayrulewidth}
AlphaGan-BG\_M &\textbf{0.0001}   &\textbf{0.001}   &\textbf{0.0006}   &0.014   &0.059   &0.014   &0.0008   &0.014   &0.001   &0.027   &0.053   &0.019   &0.0002   &\textbf{0.001}   &0.0007   \\
AlphaGan-BG\_H &\textbf{0.0002}   &\textbf{0.010}   &\textbf{0.0008}   &0.0006   &0.028   &0.001   &0.0003   &0.017   &\textbf{0.0008}   &0.001   &0.031   &0.002   &0.0003   &0.011   &\textbf{0.0008}   \\ \Xhline{2\arrayrulewidth}
\end{tabular}
\end{adjustbox}
\captionof{table}{MSE Metric - Performance of the AlphaGan-BG models using different background reconstruction methods. \textit{A: Alex, C: Castle} and \textit{D: Dimitriy}. The best average value per model for each animation across all reconstruction method is emboldened.}\label{tabcomparisonvideomse}
\end{center}

\begin{center}
\begin{adjustbox}{width=\textwidth}
\begin{tabular}{l|c|c|c|c|c|c|c|c|c|c|c|c|c|c|c}
\Xhline{2\arrayrulewidth}
 &
  \multicolumn{3}{c|}{W\textbackslash Artifact} &
  \multicolumn{3}{c|}{DVI \cite{Kim2019deepinpainting}} &
  \multicolumn{3}{c|}{DFGI \cite{Xu2019flowinpainting}} &
  \multicolumn{3}{c|}{GVI \cite{chang2019free}} &
  \multicolumn{3}{c}{FVC \cite{Afifi2014fastvideocomp}} \\ \hline
               & A & C & D & A & C & D & A & C & D & A & C & D & A & C & D \\ \Xhline{2\arrayrulewidth}
AlphaGan-BG\_M &\textbf{0.365}   &\textbf{5.779}   &\textbf{2.51}   &12.52   &123.945   &21.397   &1.376   &33.184   &3.931   &17.57   &110.25   &25.134   &0.582   &7.826   &2.835   \\
AlphaGan-BG\_H &\textbf{0.744}   &\textbf{67.829}   &\textbf{3.059}   &1.075   &95.806   &4.061   &0.766   &75.408   &3.099   &2.111   &97.737   &6.213   &0.762   &68.364   &3.098   \\ \Xhline{2\arrayrulewidth}
\end{tabular}
\end{adjustbox}
\captionof{table}{GRAD Metric - Performance of the AlphaGan-BG models using different background reconstruction methods. \textit{A: Alex, C: Castle} and \textit{D: Dimitriy}. The best average value per model for each animation across all reconstruction method is emboldened.}\label{tabcomparisonvideograd}
\end{center}

\begin{center}
\begin{adjustbox}{width=\textwidth}
\begin{tabular}{l|c|c|c|c|c|c|c|c|c|c|c|c|c|c|c}
\Xhline{2\arrayrulewidth}
 &
  \multicolumn{3}{c|}{W\textbackslash Artifact} &
  \multicolumn{3}{c|}{DVI \cite{Kim2019deepinpainting}} &
  \multicolumn{3}{c|}{DFGI \cite{Xu2019flowinpainting}} &
  \multicolumn{3}{c|}{GVI \cite{chang2019free}} &
  \multicolumn{3}{c}{FVC \cite{Afifi2014fastvideocomp}} \\ \hline
               & A & C & D & A & C & D & A & C & D & A & C & D & A & C & D \\ \Xhline{2\arrayrulewidth}
AlphaGan-BG\_M &\textbf{0.457}   &\textbf{7.77}   &\textbf{1.562}   &9.78   &100.104   &11.932   &1.21   &33.173   &2.131   &17.066   &93.838   &15.254   &0.624   &9.208   &1.68   \\
AlphaGan-BG\_H &\textbf{0.801}   &\textbf{27.864}   &\textbf{1.707}   &1.252   &55.5   &2.085   &0.818   &37.95   &1.725   &2.162   &58.965   &2.948   &0.825   &28.191   &1.719   \\ \Xhline{2\arrayrulewidth}
\end{tabular}
\end{adjustbox}
\captionof{table}{CONN Metric - Performance of the AlphaGan-BG models using different background reconstruction methods. \textit{A: Alex, C: Castle} and \textit{D: Dimitriy}. The best average value per model for each animation across all reconstruction method is emboldened.}\label{tabcomparisonvideoconn}
\end{center}

\section{Conclusion}
In this paper we proposed an approach inspired by a state of the art GAN model to validate the idea of using background information as part of the alpha matting process. The proposed approach utilises an encoder-decoder structure as generator and PacthGAN as discriminator. The input to the network consists of 7 channels, including RGB image, RGB background information and the trimap. The preliminary results of the experiments and evaluations against the benchmarked methods indicate the validity of the core idea in this research. Using the full or partial background information, AlphaGan-BG demonstrated a superior performance against the studied methods.
In the future work, we would like to train and analyse the performance of AlphaGan-BG on synthetic data. The background reconstruction process is another exciting aspect of this research that requires more investigation. The current performance of the models is achieved by simulating the reconstruction artifacts. However, we believe that AlphaGan-BG can obtain a higher accuracy if trained on a specific background reconstruction method with consistent noise and artifact pattern.

\printbibliography
\newpage
\section*{Appendix 1: Visual Results on Adobe Matting dataset}
\begin{center}
\includegraphics[width=\textwidth,height=\textheight,keepaspectratio]{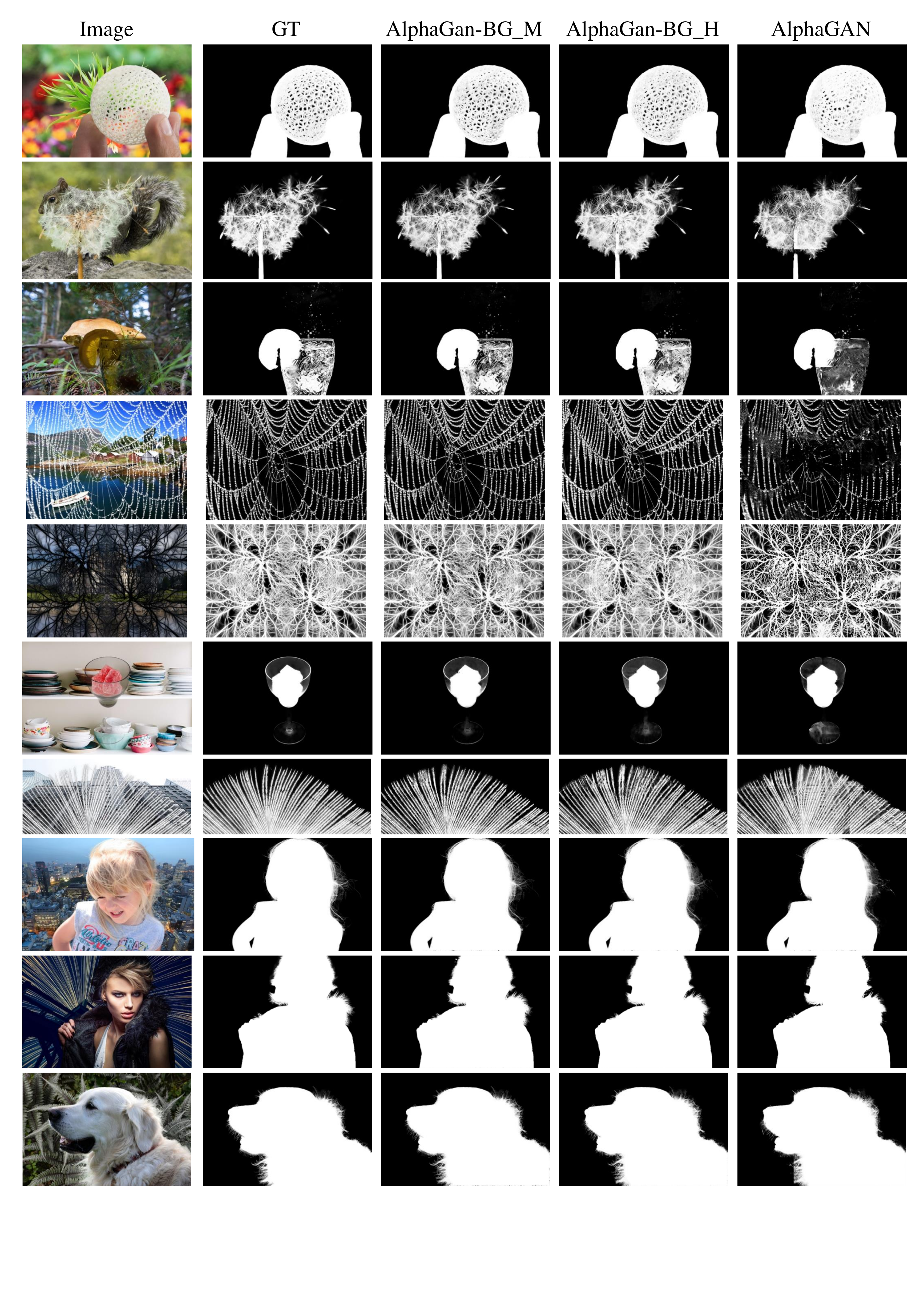}
\end{center}
\newpage
\section*{Appendix 2: Visual Results on Video Matting dataset - A Frame from Alex Video Sequence}
\begin{center}
\includegraphics[width=\textwidth,height=\textheight,keepaspectratio]{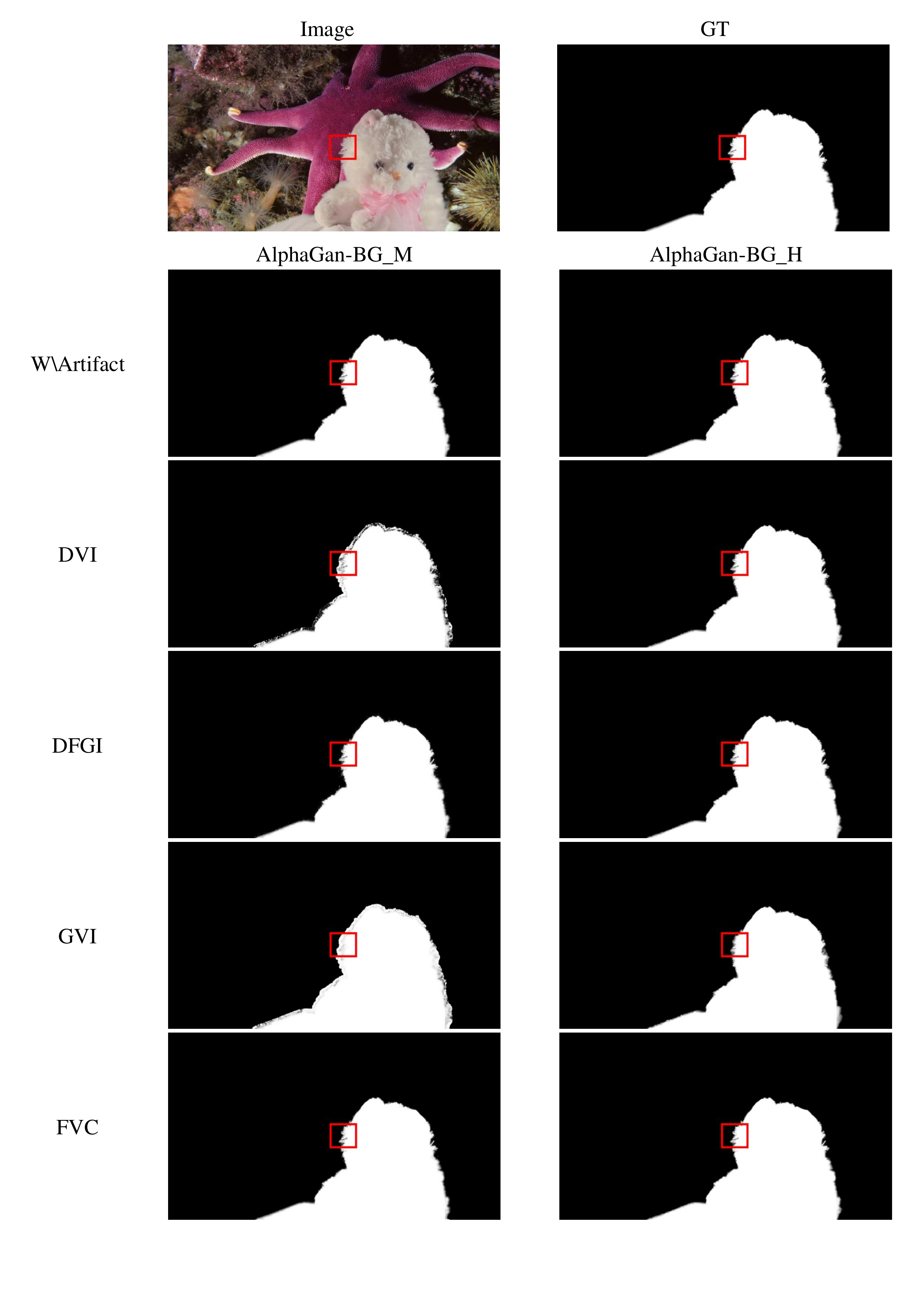}
\end{center}
\newpage
\section*{Appendix 2: Visual Results on Video Matting dataset - A Frame from Castle Video Sequence}
\begin{center}
\includegraphics[width=\textwidth,height=\textheight,keepaspectratio]{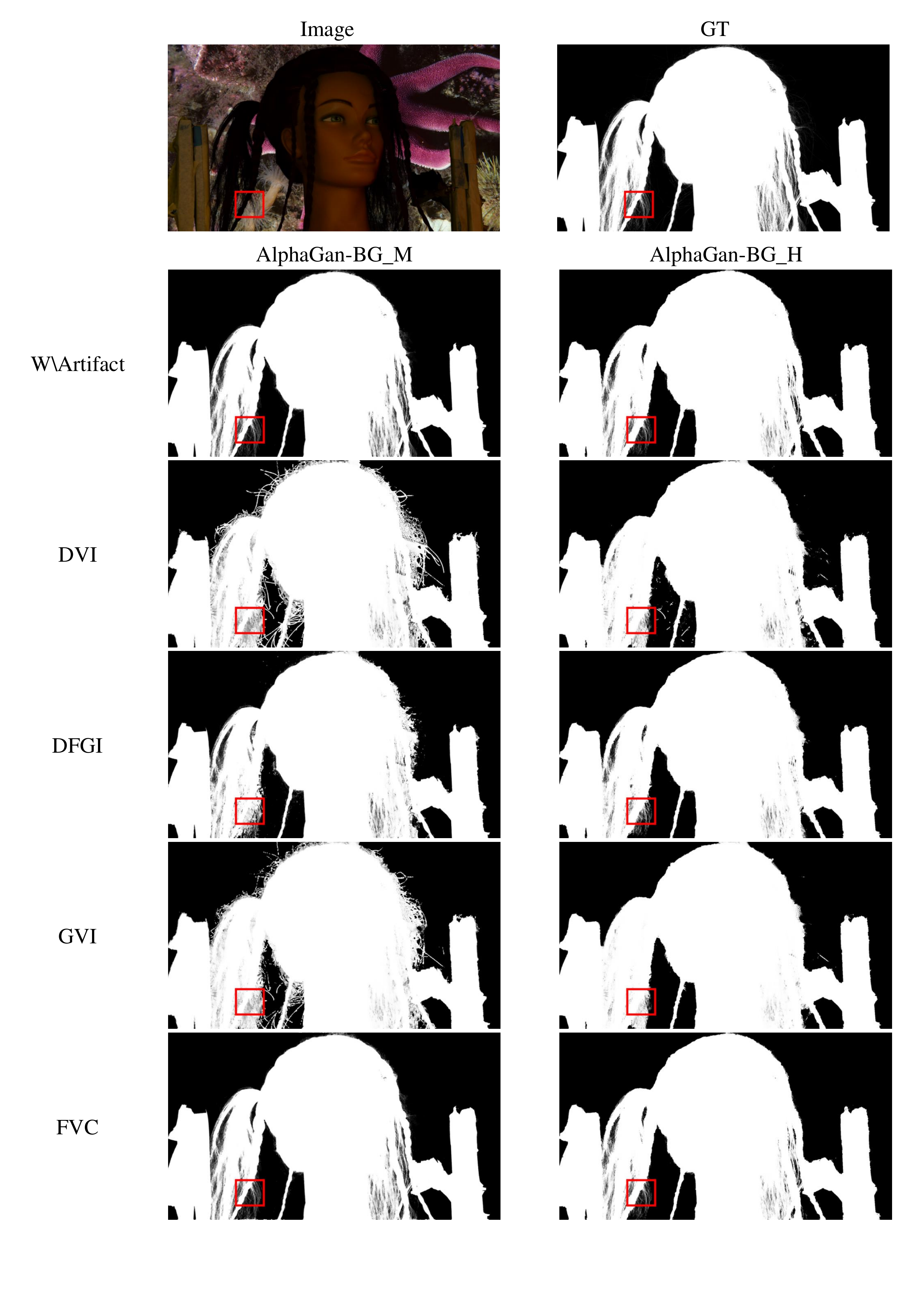}
\end{center}
\newpage
\section*{Appendix 2: Visual Results on Video Matting dataset - A Frame from Dmitriy Video Sequence}
\begin{center}
\includegraphics[width=\textwidth,height=\textheight,keepaspectratio]{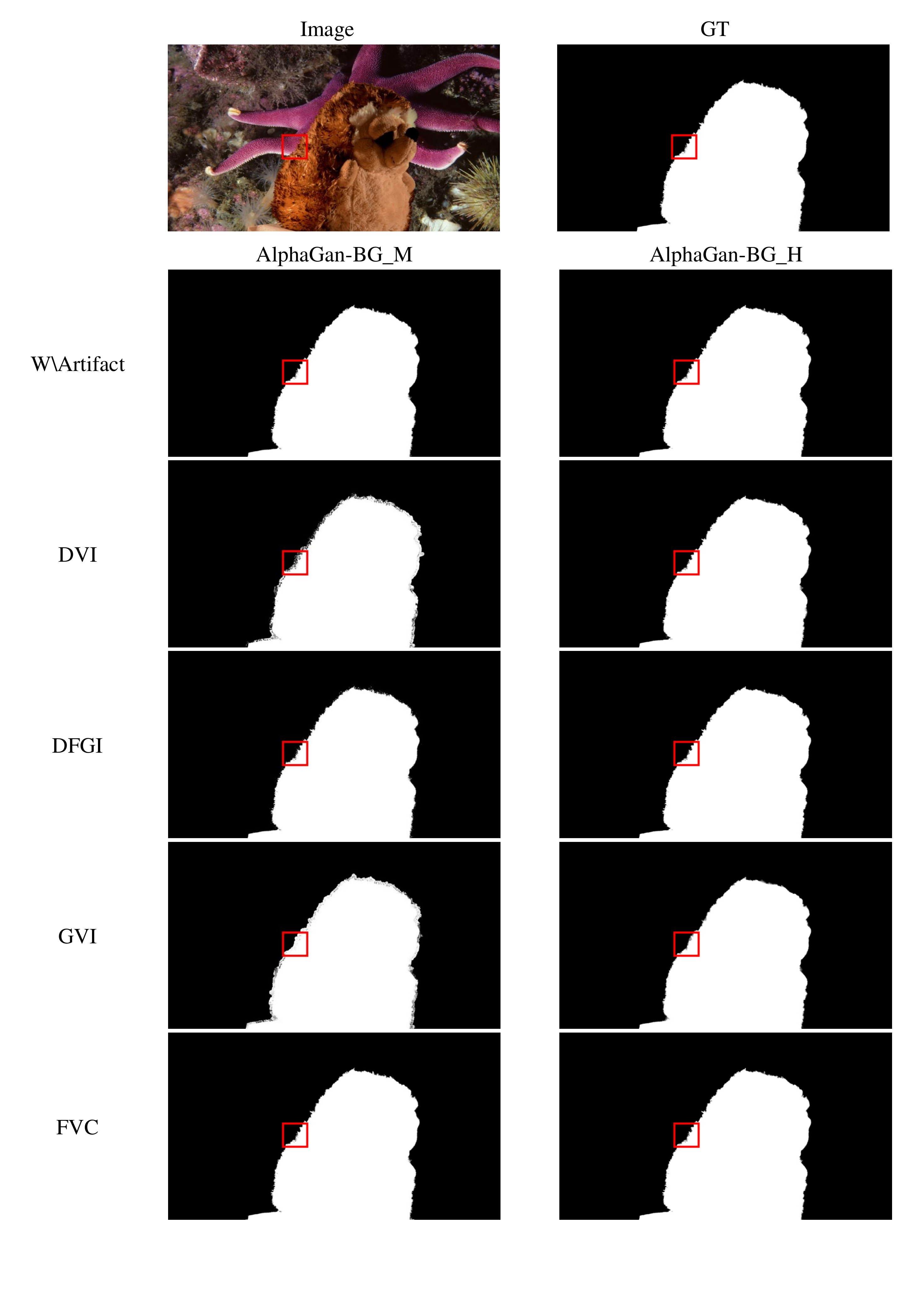}
\end{center}
\end{document}